\documentclass[sts]{imsart}

\RequirePackage{amsthm,amsmath,amsfonts,amssymb}
\RequirePackage[numbers,sort&compress]{natbib}
\RequirePackage[colorlinks,citecolor=blue,urlcolor=blue]{hyperref}
\RequirePackage{graphicx}

\startlocaldefs
\theoremstyle{plain}

\theoremstyle{definition}

\RequirePackage{amsthm,amsmath,amsfonts,amssymb}
\RequirePackage[numbers,sort&compress]{natbib}
\RequirePackage[colorlinks,citecolor=blue,urlcolor=blue]{hyperref}
\usepackage{graphicx}
\usepackage{enumerate}
\usepackage{natbib}
\usepackage{url} 
\usepackage{scalerel}
\usepackage{mathtools}
\usepackage{bbding}
\usepackage{tabularx}

\allowdisplaybreaks[4]
\usepackage{float}
\usepackage{mathrsfs}
\usepackage{bm}
\usepackage{colonequals}
\usepackage{graphicx}
\usepackage{caption}
\usepackage{enumerate}
\usepackage{cmll}
\usepackage{amsthm}
\usepackage{subcaption}
\usepackage{threeparttable}
\usepackage{booktabs}
\usepackage{multirow}
\usepackage{makecell}
\usepackage[table,dvipsnames]{xcolor}
\usepackage{tikz}
\usetikzlibrary{positioning,arrows.meta,quotes}
\usetikzlibrary{shapes,snakes}
\usetikzlibrary{bayesnet}
\tikzset{>=latex}
\tikzstyle{plate caption} = [caption, node distance=0, inner sep=0pt,
below left=5pt and 0pt of #1.south]
\usepackage{algorithm}
\usepackage{algpseudocode}

\startlocaldefs

\makeatletter

\makeatletter

\newcommand{\cO}{\mathcal{O}}

\newcommand{\hY}{\widehat{Y}}

\newcommand{\hX}{\widehat{X}}

\newcommand{\T}{\rm T}

\newcommand{\cT}{\scaleto{\mathcal{T}\mathstrut}{5.5pt}}

\newcommand{\blue}{\color{blue}}

\newcommand{\specialcell}[2][c]{%
  \begin{tabular}[#1]{@{}l@{}}#2\end{tabular}}

\graphicspath{{./fig/}}
\endlocaldefs

\begin{document}

\begin{frontmatter}
\title{Harnessing Synthetic Data  from Generative AI for  Statistical Inference}  
\runtitle{Synthetic Data for Statistical Inference}
\bigskip
\begin{aug}
\author[A]{\fnms{Ahmad}~\snm{Abdel-Azim}\ead[label=e1]{ahmad\_abdelazim@g.harvard.edu}\orcid{0000-0003-1268-3366
}
},
\author[B]{\fnms{Ruoyu}~\snm{Wang}\ead[label=e2]{ruoyuwang@hsph.harvard.edu}\orcid{0000-0002-4561-2954
}}
\and
\author[C]{\fnms{Xihong}~\snm{Lin}\ead[label=e3]{xlin@hsph.harvard.edu}\orcid{0000-0001-7067-7752}}

\thankstext{t1}{The first two authors contributed equally. Alphabetical order.}

\address[A]{Ahmad Abdel-Azim is Ph.D. Candidate, Department of Biostatistics,
Harvard T.H. Chan School of Public Health, Boston, Massachusetts 02115, USA\printead[presep={\ }]{e1}.}

\address[B]{Ruoyu Wang is Postdoctoral Fellow, Department of Biostatistics,
Harvard T.H. Chan School of Public Health, Boston, Massachusetts 02115, USA\printead[presep={\ }]{e2}.}

\address[C]{Xihong Lin is Professor of Biostatistics and  Professor of Statistics,
Harvard University, Boston, Massachusetts 02115, USA\printead[presep={\ }]{e3}.}

\end{aug}

\begin{abstract}
The emergence of generative AI models has dramatically expanded the availability and use of synthetic data across scientific, industrial, and policy domains. While these developments open new possibilities for data analysis, they also raise fundamental statistical questions about when synthetic data can be used in a valid, reliable, and principled manner. This paper reviews the current landscape of synthetic data generation and use from a statistical perspective, with the goal of clarifying the assumptions under which synthetic data can meaningfully support downstream discovery, inference, and prediction. We survey major classes of modern generative models, their intended use cases, and the benefits they offer, while also highlighting their limitations and characteristic failure modes. We additionally examine common pitfalls that arise when synthetic data are treated as surrogates for real observations, including biases from model misspecification, attenuated uncertainty, and difficulties in generalization. Building on these insights, we discuss emerging frameworks for the principled use of synthetic data. We conclude with practical recommendations, open problems, and cautions intended to guide both method developers and applied researchers.
\end{abstract}

{\blue 
}
\end{frontmatter}

\section{Introduction}
 
Recent breakthroughs in deep generative modeling and the explosive growth of generative artificial intelligence (AI) have dramatically expanded what is possible with synthetic data, and this has garnered attention across statistics, machine learning, and applied domains. Large language models (LLMs) \cite{brown2020language}, diffusion models \cite{ho2020denoising}, and other breakthrough models have moved beyond research curiosities to integrated tools used actively in scientific discovery, healthcare, autonomous systems, and even personal daily use. Modern generators are capable of producing high-fidelity synthetic observations in high dimensions. These capabilities have enabled synthetic data to be used not only for privacy-preserving data release, as originally proposed \cite{rubin1993statistical}, but also for a wide range of exciting applications \cite{mccaw2024synthetic,bing2022conditional, an2023deep, zhu2017unpaired, xu2019modeling}.
Yet, this rapid evolution of generative modeling has outpaced our understanding of when and how synthetic data can reliably support downstream statistical inference and scientific discovery. 

Synthetic data refer to generated artificial data that have mimics  the distribution of the original data \cite{chen2021synthetic}. We consider ``data-driven'' synthetic data in particular here \cite{van2023synthetic},  especially high-dimensional data; this focuses us on data generated from modern generative AI models, rather than traditional synthetic datasets generated by user-specified mathematical equations or simulations that are typically oversimplified and are difficult to specify for high-dimensional data \cite{zhu2024generating, jarmin2014expanding, predhumeau2023synthetic}.

Traditionally, synthetic data were developed to facilitate downstream analyses while reducing disclosure risk and satisfying the regulatory privacy constraints that may restrict access to sensitive original data (e.g., medical records) \cite{kaloskampis2020synthetic, van2023synthetic, zhang2020ensuring, hyrup2024sharing, ho2021dp}.
Recently, however, with the introduction of improved deep learning-based generators, there is an increasing interest in leveraging synthetic data to improve upon real data, for example to improve fairness \cite{van2021decaf, xu2018fairgan, xu2019achieving}, augment the dataset size to increase inferential power \cite{mccaw2024synthetic, miao2024valid, antoniou2017data, bing2022conditional, das2022conditional}, create data for new domains \cite{wang2019learning, yoon2018radialgan}, and generate synthetic tasks for in-context learning \citep{mullertransformers,hollmanntabpfn,hollmann2025accurate}.  In the modern age of deep learning and AI, the predominant focus has been on developing powerful generative models that can approximate sampling from complex and high-dimensional data distributions. Approaches such as generative adversarial networks (GANs) \cite{goodfellow2014generative} and variational autoencoders (VAEs) \cite{kingma2013auto} have evolved into scalable transformer-based LLMs and score-based diffusion models \cite{vaswani2017attention, brown2020language, ho2020denoising, song2020score}, enabling the generation of realistic images, text, and multimodal content and spurring new applications for synthetic data across several domains, from augmenting medical imaging datasets to improve diagnostic model performance \cite{lo2021medical} to generating synthetic samples to alleviate class imbalance in fraud detection systems \cite{charitou2021synthetic}. 

Such advances in generative AI have led to vast quantities of synthetic data being produced and used across several industry domains. For example, synthetic data permeates LLM training pipelines. While its utility has been fairly established in post-training stages such as instruction-tuning and alignment where objectives are specified and data are scarce \cite{li2023self, ge2024scaling}, its uses and effects in pre-training are poorly understood \cite{liu2024best}. Naive recursive training of LLMs solely on generated outputs from earlier generations, namely treating synthetic data as real without careful control, has been shown to lead to ``model collapse'', where learned models progressively lose diversity and misrepresent tails of the original data distribution \cite{shumailov2024ai}. 

Failure modes like this emphasize the need for principled frameworks and highlight both the promise and risk of synthetic data usage. Despite the breadth of work in generative modeling, it is critical to also understand when synthetic data can safely support downstream analytic tasks and what methodological guardrails can be leveraged to make synthetic data use statistically valid \cite{van2023synthetic}.  Naively combining real observed data with synthetic data by treating synthetic data the same as the observed data in statistical analysis can lead to invalid inference and bias \cite{decruyenaere2023real, drechsler202430}. 

Furthermore, generative models are often misspecified in practice. 
Synthetic samples drawn from misspecified generative models
can systematically misrepresent key features of the target distribution. To ensure robustness to model misspecification, it is critical to understand how synthesis error and model misspecification propagate through inferential and predictive workflows.

This article reviews the generation and use of synthetic data from a statistical perspective. 
Unlike reviews of deep generative modeling (e.g., \cite{bond2021deep}) or surveys of synthetic data use cases (e.g., \cite{van2024synthetic}),   our focus is on the statistical conditions and methodological frameworks under which synthetic data enable valid and powerful downstream analysis, inference, and prediction, particularly when generative models used to produce synthetic data are misspecified. We demonstrate that integrating statistical principles with generative AI models can empower trustworthy scientific discovery through the effective use of synthetic data.

We first give an organized framework for the main motivations for generating synthetic data, particularly with the advent of modern generative models capable of producing high-fidelity samples.  We then briefly survey the major generative model classes, focusing in particular on their assumptions and appropriate use cases. Importantly, we emphasize the properties that are critical for their downstream use and distill the statistical challenges that govern downstream utility, such as issues of model misspecification and the uncertainty of synthetic data. We organize and compare statistically principled paradigms for using synthetic data, delineating the assumptions under which each can provide plausible guarantees.  We provide a few practical data examples to illustrate applications that leverage synthetic data. Finally, we highlight practical considerations for synthetic data usage and identify open problems where new statistical theory and methodology are needed to leverage high-fidelity generations for reliable scientific discovery.

\section{Synthetic data generation}

\subsection{Why synthetic data?}
\label{sec:whySynth}

Despite only being first proposed as a privacy preserving measure \cite{rubin1993statistical}, there exist several motivations for generating synthetic data, especially as modern generators become increasingly powerful. From a statistical perspective, synthetic data generation can be framed as selecting a sampling distribution and a use protocol tailored to a downstream objective.

Let $X$ and $Y$ be the predictor and label, respectively. Let the observed dataset $\mathcal O$ be $n$ observations drawn from an unknown training distribution $P$. When a target population distinct from the training distribution is relevant, write it as $P_{\cT}$; synthetic data are drawn from a sampling distribution $Q$. A generative procedure trained on $\mathcal O$ or some external datasets learns an estimated distribution $\widehat P_{\theta}$, which depends on an unknown parameter $\theta$,
inducing a synthetic sampling distribution $Q$ (often $Q= \widehat P_\theta$), from which we draw a synthetic dataset $\mathcal S$.
Downstream goals target a functional $\beta(P_{\cT})$ (e.g., a causal estimand or regression parameter) or a predictive object $m_{P_{\cT}}$ (e.g., a risk score or conditional mean), with $P_{\cT}=P$ in the absence of distribution shift.

Specifically, suppose $\mathcal O=\{Z_i\}_{i=1}^n \sim Z = (X,Y)$ sampled from a training distribution $P$ are observed in the real dataset, along with auxiliary variables $\{A_{i}\}_{i = 1}^{n} \sim A$ (e.g., attribute/domain label/time/intervention). The auxiliary variable is not of primary interest but can be useful in guiding the generation of synthetic data, such as when the target is to generate $(X, Y)$ under a specific intervention. The auxiliary variable $A$ plays a central role in conditional synthetic data generation. In some settings, like cases of partially missing or coarsened data, $A$ is observed, and synthetic data are generated or imputed conditional on the observed $A$. In other settings, such as in cases of fully synthetic data generation, auxiliary variables are unobserved and must first be drawn from a specified or estimated distribution, after which data are generated conditional on the sampled values of $A$. A generative model trained on $\mathcal O$ induces an estimated distribution $\widehat P_\theta$ and the target sampling distribution $Q$, from which synthetic data  $\mathcal S$, which include $\widehat{X}$ and/or $\widehat{Y}$ are drawn.  We organize the main motivations for generating synthetic data into five broad categories.

Table \ref{tab:SDGmotivations} summarizes five common motivations for generating synthetic data from a statistical perspective. 
We organize the different motivations in a simple framework; each synthetic data aim is characterized by (i) the target sampling distribution $Q$ from which synthetic records $\mathcal S$ are drawn, and (ii) the \emph{access pattern} specifying how an analyst interacts with $(\mathcal O,\mathcal S)$ downstream. These two axes separate settings where $Q$ is intended as a plug-in approximation to the training distribution $P$ from settings where $Q$ is deliberately conditional (conditional data augmentation or trajectory completion), shifted toward a different target population (domain transfer with $Q\approx P_{\cT}$), or constrained to satisfy an external criterion (privacy-preserving release or fairness with $Q^\star$).

The focus here is not a comprehensive survey of applications, for which good reviews exist (e.g., see \cite{drechsler202430}), but rather the statistical aims that motivate generating synthetic data. We discuss below the five synthetic data generation settings in detail.

\begin{table*}[]
\scriptsize
\caption{Different motivations for generating synthetic data}\label{tab:SDGmotivations}
\begin{tabular}{@{}m{2.6cm} m{3.3cm} m{3.5cm} m{3.5cm} m{2.5cm}@{}}
\toprule
\multicolumn{1}{c}{\textbf{Motivation}} &
  \multicolumn{1}{c}{\textbf{Target sampling distribution $Q$}} &
  \multicolumn{1}{c}{\textbf{Intended objective}} &
  \multicolumn{1}{c}{\textbf{Access pattern for $(\mathcal O,\mathcal S)$}} &
  \multicolumn{1}{c}{\textbf{Example methods}} \\ \midrule
  \specialcell[b]{\textbf{Privacy-preserving} \\ \textbf{release}} &
  \specialcell[b]{Single release: $Q=\widehat P_\theta \approx P$; \\Multiple releases: \\$Q(z)=\int \widehat   P_\theta(z)\,p(\theta\mid \mathcal O)\,d\theta$}  &
  Enable analysis without   releasing individual-level records; propagate synthesis uncertainty or   enforce a formal privacy guarantee. &
  Use $\mathcal S^{(1)},\dots,\mathcal S^{(M)}$ only (often multiple releases) &
  MI \cite{reiter2003inference, raghunathan2003multiple, drechsler2011synthetic}; DP-SGD \cite{abadi2016deep}; PATE \cite{papernot2016semi}; PrivBayes \cite{zhang2017privbayes}; PATE-GAN \cite{jordon2018pate}
   \\ \midrule
\textbf{Data augmentation} &
  \specialcell[b]{Unconditional: $Q = \widehat P_\theta \approx P$\\Conditional: \\ $Q(\cdot)=\widehat P_\theta(\cdot\mid A=a)$, or \\$\widehat P_\theta(\cdot\mid Y=1)$} &
  Increase sample size and diversity for downstream learning (e.g., improved predictive performance, model tuning) &
  Use $\mathcal O\cup\mathcal S$, with potential calibration or weighting steps & TabDDPM \cite{kotelnikov2023tabddpm}; SMOTE \cite{chawla2002smote}; CTAB-GAN \cite{zhao2021ctab}; CTGAN \cite{xu2019modeling}; CoDSA \cite{tian2025conditional}
   \\ \midrule
\textbf{Fairness} &
  \specialcell[b]{Constrained: \\ $Q^\star\in\arg\min_{Q\in\mathcal Q} D(Q,\widehat P_\theta)$ \\ s.t. $\mathrm{Fair}(Q)\ge\tau$} &
  Modify the training distribution   to satisfy a fairness criterion while preserving utility. &
  Typically train/analyze on   $\mathcal O\cup \mathcal S$, or use $\mathcal S$ to reduce bias; protocols vary by fairness goal. & FairGAN \cite{xu2018fairgan}; DECAF \cite{van2021decaf}; TabFairGAN \cite{rajabi2022tabfairgan}
   \\ \midrule
\textbf{Domain transfer} &
  \specialcell[b]{Target-domain synthesis: \\ $Q\approx P_{\cT}$ using training data \\ $P \neq P_{\cT}$; covariate shift: \\$P_{\cT}(Y\mid X) = P(Y\mid X)$} &
  Improve target performance under shift by evaluating on data matching the target environment. &
  Use synthetic transfer samples $\mathcal S \sim Q\approx P_{\cT}$ (with $\mathcal O\sim P$) to approximate training under $P_{\cT}$ & IWCV \cite{sugiyama2007covariate}; RadialGAN \cite{yoon2018radialgan}; Optimal transport \cite{courty2016optimal, courty2017joint, balaji2020robust, nguyen2021most}
   \\ \midrule
   \specialcell[b]{\textbf{Missing data/trajectory}\\  
   \textbf{completion}} &
  \specialcell[b]{Conditional completion: \\ $Q=\widehat P_\theta(Z_{\text{miss}}\mid Z_{\text{obs}},A)$; \\ Forecasting (special case): \\$Q=\widehat P_\theta(Z_{t_0+1:T}\mid Z_{1:t_0},A)$}&
  Fill in missing data segments (e.g., forecast future) and/or generate plausible futures (e.g., digital   twins). &
  Sample missing data (trajectories) conditioned  on observed data (prefix); used for imputation, simulation, forecasting. & CSDI \cite{tashiro2021csdi}; TimeGAN \cite{yoon2019time}; BRITS \cite{cao2018brits}; DT-GPT \cite{makarov2025large}
   \\ \bottomrule
\end{tabular}
\end{table*}

\subsubsection{Privacy-preserving release} is often regarded as the original motivation for synthetic data generation when individual-level records in $\mathcal O$ cannot be shared.  The defining access pattern is that external analysts typically only receive (often multiple) synthetic releases $\mathcal S^{(1)},\dots,\mathcal S^{(M)}$ and do not access $\mathcal O$ directly \cite{rubin1993statistical, reiter2003inference, raghunathan2003multiple}. The target sampling distribution $Q$ is thus chosen to approximate the training distribution $P$ while meeting an explicit privacy constraint so that downstream analyses can proceed using $\mathcal S$ alone.

A few widely used frameworks are examples of this idea.   In its simplest form, a single synthetic release draws data from a plug-in estimate of the training distribution so that \(Q = \widehat P_\theta \approx P\).
In the classical multiple-imputation (MI) formulation, the generative model parameters are assumed to be random rather than fixed, so the synthetic sampling law is defined by averaging over the model parameter posterior distributions. Synthesis uncertainty is propagated by integrating over generative model parameter uncertainty, and the resulting target distribution is the posterior predictive mixture \cite{rubin1987},
\[
Q(z) = \int \widehat P_\theta(z) p(\theta | \mathcal O) d\theta,
\]
where the learned generative model $\widehat P_\theta(z)$ here represents the model's conditional predictive distribution for the quantity being synthesized, namely the imputation distribution for unobserved values in the Bayesian MI perspective. Note that dependence of $\widehat P_\theta$ on the observed data $\mathcal O$ is left implicit in the notation above once the parameters of the generative model are estimated and fixed.
From a Bayesian perspective, each synthetic release is generated by first drawing $\theta^{(m)} \sim p(\theta \mid \mathcal O)$ and then sampling from $\widehat P_{\theta^{(m)}}$. 
Downstream uncertainty reflects both sampling variability and uncertainty in the generative model parameters \cite{reiter2003inference, raghunathan2003multiple, drechsler2011synthetic}. 

In modern differential privacy (DP) formulations, $Q$ is regarded as the output law of a randomized mechanism $M$, which takes a dataset as input and outputs an element in an output space $\mathcal Y$, satisfying DP constraints. In particular, for $\varepsilon\ge 0$ and $\delta\in[0,1]$, $M$ is $(\varepsilon,\delta)$-differentially private if for all adjacent datasets $\mathcal O,\mathcal O'$ (i.e., datasets that differ by one individual record) and all measurable $B\subseteq \mathcal Y$,
\[
\mathbb P \left(M(\mathcal O)\in B\right) \le e^{\varepsilon}\mathbb P \left(M(\mathcal O')\in B\right) + \delta,
\]
where the probabilities are taken over the randomness of $M$ \cite{dwork2014algorithmic}. Critically, this constraint forces deliberate random perturbation of the data-generating process, and thus $Q$ is, by design, shifted away from  $\widehat P_\theta$, namely the plug-in approximation to the data-generating distribution that would have been learned without the DP constraints. 

For example, in the canonical differentially private mean estimation problem, standard techniques first clip sample values to a bounded range to control sensitivity and then add noise \cite{kamath2020private}. While this clipping satisfies the privacy constraint, it introduces a non-vanishing statistical bias that cannot be removed without violating differential privacy. Allowing weaker privacy guarantees, as tuned by $(\varepsilon,\delta)$, reduces this bias \cite{kamath2025bias}. The required stability to single record perturbations limits fidelity of the induced synthetic distribution $Q$ relative to unconstrained estimators, making the privacy-utility tradeoff explicit and unavoidable in many cases \cite{dwork2014algorithmic, geng2014optimal}.

In practice, this perspective underlies deep learning approaches such as DP-SGD \cite{abadi2016deep}, which are used to train private generators, and teacher ensemble approaches such as PATE \cite{papernot2016semi}, which can be combined with generative modeling for private synthesis (e.g., PATE-GAN \cite{jordon2018pate}). DP synthetic release also includes structured mechanisms such as PrivBayes \cite{zhang2017privbayes}, which privately estimates a graphical model and samples synthetic records from the resulting DP approximation.

\subsubsection{Data augmentation}  
covers settings where individual-level records in $\mathcal O$ remain available, but the sample size or diversity is a limiting factor for a downstream task. In contrast to privacy-preserving release, the defining access pattern here is joint access to both real and synthetic data. The observed data $\mathcal O$ remains available and is merged with $\mathcal S$ to increase effective sample size for downstream procedures (e.g., to stabilize training, support model selection, or improve statistical power and finite-sample behavior) \cite{tian2025conditional, chang2019kernel}. In its simplest form, the target sampling distribution is a plug-in approximation $Q=\widehat P_\theta\approx P$, with the objective of drawing additional samples that are consistent with the structure learned from $\mathcal O$.
Conceptually, this can be viewed as interpolating along a data manifold learned from $\mathcal O$, making it distinct from bootstrapping, which resamples only from the discrete empirical distribution supported on the observed records.

Consider for example the supervised learning objective of training a stable predictor of $Y$ from $X$ with limited labeled data (e.g., a modest clinical cohort). A data expansion pipeline fits $\widehat P_\theta$ to the joint law of $(X,Y)$ using (a model-building subset of) $\mathcal O$, samples $\mathcal S=\{\widehat X_j,\widehat Y_j\}_{j=1}^N\sim Q = \widehat P_\theta$, and trains or tunes downstream procedures on $\mathcal O\cup\mathcal S$. Leveraging such synthetic data while maintaining valid inference is a key consideration discussed in Section \ref{sec:use}.

In the context of supervised learning with limited labeled data, this motivation has driven the development of generators for tabular data, where mixed variable types, skewed marginals, and complex dependencies make high fidelity synthetic data sampling challenging. Representative tabular GAN work explicitly targets challenges like mixed continuous and categorical structure, long-tailed continuous variables, and imbalanced categories; failure to capture these dependencies produce synthetic artifacts that degrade downstream utility rather than expanding effective sample size \cite{zhao2021ctab}. 

More recently, diffusion-based tabular generators have been proposed showing that denoising diffusion models can be adapted to tabular distributions and even outperform earlier GAN/VAE baselines on common utility benchmarks \cite{kotelnikov2023tabddpm}. Further, since generators can overfit when the data size is limited leading to poor performance, class-conditional energy-based synthesis have been proposed to stabilize generation, reporting consistent gains from augmentation on small tabular classification datasets relative to using $\mathcal O$ alone \cite{margeloiu2024tabebm}.

 While overall sample size is sometimes the limiting factor for downstream performance, oftentimes, it is scarcity in a specific region of the joint law, such as a rare class label, a rare outcome, or an underrepresented subpopulation that limits downstream performance or efficiency \cite{tian2025conditional, chawla2002smote}. In such cases, the target sampling distribution $Q$ is specified as a conditional law so synthetic sampling targets a designated region of the data space rather than approximating $P$ marginally; for example, one could choose,
\[
Q(\cdot)=\widehat P_\theta(\cdot\mid A=a)\quad\text{or}\quad  Q(\cdot)=\widehat P_\theta(\cdot\mid Y=1),
\]
for a data attribute $a$. Note that conditioning on $Y=1$ refers to rare event oversampling in supervised learning settings; for example, in a clinical cohort with $Y\in\{0,1\}$ indicating a rare adverse event, generating additional $(X,Y)$ pairs with $Y=1$ increases the number of cases for improved model training and evaluation \cite{chawla2002smote}. Recognize that the access pattern here is non-restrictive as $\mathcal O$ remains available and synthetic samples $\mathcal S\sim Q$ are added to mitigate imbalance rather than replace the dataset.

Conditional data augmentation has motivated both classical and modern conditional generators. SMOTE is a canonical baseline that constructs synthetic minority-class feature vectors by interpolating between nearby minority examples \cite{chawla2002smote}. In the modern age of deep learning, conditional GANs have been proposed to better sample $X$ conditional on a label or attribute (e.g., $Y$ or $A$) \cite{mirza2014conditional}. In tabular settings, CTGAN, for example, samples a conditioning value during training (``training-by-sampling'') so that rare categories in discrete columns are explicitly represented in the learned conditional generator \cite{xu2019modeling}. More recently, CoDSA leverages more modern generative models, like diffusion models, and frames targeted synthesis as allocating synthetic mass to under-sampled or otherwise user-specified regions via conditional generation,  e.g., using conditional diffusion models and fine-tuning \cite{rombach2022high, tian2025conditional}. Further, a theoretical framework is proposed to quantify the statistical gains to be expected as a function of synthetic sample size and region allocation; this is supported by empirical comparisons to non-targeted augmentation baselines \cite{tian2025conditional}.

\subsubsection{Fairness.} 

In many applications, the goal is not to approximate the training distribution $P$ as closely as possible, but rather to construct synthetic data that modifies the effective training distribution so downstream decisions satisfy a specified fairness criterion with respect to a protected sensitive attribute $A$ (e.g., sex, race). A canonical example is risk scoring for loan approval, where $Y$ denotes an outcome (e.g., ``default on a loan''). When historical data encode structural inequities, a predictor trained directly on $\mathcal O\sim P$ can inherit systematic disparities across groups. Fairness goals formalize which disparities are deemed unacceptable such as parity constraints on prediction rates across groups (demographic parity) or parity of true positive rates across groups \cite{hardt2016equality}.

Fairness-motivated synthesis is distinguished by its target sampling law. Rather than seeking $Q\approx P$, the target is a constrained distribution $Q^\star\neq P$ that trades off fidelity to the learned data distribution against a fairness constraint:
\[
Q^\star \in \arg\min_{Q\in\mathcal Q} D \left(Q,\widehat P_\theta\right)
\quad\text{s.t.}\quad \mathrm{Fair}(Q)\ge \tau,
\]
where $D$ measures discrepancy from $\widehat P_\theta$ (commonly chosen as an adversarial discrepancy or a divergence proxy) and $\mathrm{Fair}(Q)$ encodes the chosen fairness notion \cite{verma2018fairness}. For example, fixing a downstream predictor $\widehat Y$ that is a learned function of $X$, a notion of fairness can be encoded as $\mathrm{Fair}(Q)=-\left|\mathbb P(\widehat Y=1\mid A=0)-\mathbb P(\widehat Y=1\mid A=1)\right|$ \cite{hardt2016equality, verma2018fairness}. As such, synthetic data are not a privacy-preserving or power-enhancing surrogate for $\mathcal O$, but rather a principled way to adjust the training distribution so downstream procedures better align with a fairness target, while retaining utility.

The data access pattern is typically non-restrictive as $\mathcal O$ often remains available to the analyst; synthetic data $\mathcal S\sim Q^\star$ are used to modify training and enforce fairness properties in the data distribution. Several approaches leverage adversarial objectives to perform the constrained generation. FairGAN, for example, adds an explicit fairness component to GAN training so generated samples are not only indistinguishable from real data but also less informative about the protected attribute according to the target criterion \cite{xu2018fairgan}; TabFairGAN adapts this paradigm to tabular synthesis \cite{rajabi2022tabfairgan}. Further, since fairness notions can be inherently causal (e.g., separating permissible from impermissible causal pathways from $A$ to $Y$), causally-aware generators such as DECAF \cite{van2021decaf} use an explicit structural model to synthesize data under fairness constraints imposed on mechanisms, rather than only on observed marginals.

\subsubsection{Domain transfer} refers to settings in which the target population differs from the available training population. 
In a multi-site biomedical prediction task for example, a model may be trained on a ``source'' hospital dataset but deployed at a ``target'' hospital with different covariate distributions, measurement practices, or prevalence, so that predictors and outcomes follow different joint laws \cite{steingrimsson2023transporting, guo2022evaluation}. Formally, the observed training data satisfy $\mathcal O=\{X_i,Y_i\}_{i=1}^n\sim P$, but the inferential or deployment target is a distribution $P_{\cT}\neq P$. Rather than drawing synthetic samples from $\widehat P_\theta \approx P$, the goal is to learn a target sampling distribution $Q$ that approximates the target distribution $P_{\cT}$, using only access to source observations $\mathcal O\sim P$ and typically some target information (e.g., unlabeled target covariates, or limited target labels). Here, the intended objective is to reduce the mismatch between the effective training distribution and the target environment, and downstream procedures are trained and evaluated using synthetic transfer samples $\mathcal S\sim Q$ so that learning better reflects $P_{\cT}$. For example, RadialGAN \cite{yoon2018radialgan} learns cross-dataset mappings with multiple GANs so that samples from related source datasets can be transformed into the target domain, effectively enlarging the target training set. 

A common special case is covariate shift, where the conditional outcome model is assumed stable across domains while the covariate distribution changes, 
\[
P_{\cT}(Y\mid X)=P(Y\mid X), \quad P_{\cT}(X)\neq P(X).
\] 
Under this assumption, target risk can be expressed as a reweighted source expectation, motivating importance weighting and related model-selection criteria (e.g., importance-weighted cross-validation \cite{sugiyama2007covariate}) that seek to tune predictors for $P_{\cT}$ using source-labeled data and target covariates.

Another prominent class of explicit transfer mechanisms uses optimal transport (OT) to align source and target distributions by estimating a coupling or induced mapping that moves source samples toward the target \cite{courty2016optimal, courty2017joint, balaji2020robust, nguyen2021most}. In the OT-based domain adaptation, the learned transport plan can be used to ``push'' source points such that transported samples match the target distribution. Early formulations develop OT as a practical tool for domain adaptation under distribution mismatch, and later variants incorporate label information to transport joint distributions \cite{courty2017joint} or more directly preserve predictive structure while aligning domains \cite{courty2016optimal, balaji2020robust}.

\subsubsection{Missing data/trajectory completion} 
captures missing data or sequential/ordered data settings in which synthetic generation targets unobserved components of partially observed records.  In the trajectory completion setting, denote a trajectory $Z_{1:T}=(Z_1,\dots,Z_T)$; this could be a multivariate time series, an series of events, or a sequence of tokens (e.g., in natural language settings). The observed portion can be written $Z_{\mathrm{obs}}=\{Z_t:t\in\mathcal I\}$ for an index set $\mathcal I\subset\{1,\dots,T\}$, with missing targets $Z_{\mathrm{miss}}=\{Z_t:t\notin\mathcal I\}$. The synthetic target is then a conditional law
\[
Q = \widehat P_\theta\left(Z_{\mathrm{miss}} \mid Z_{\mathrm{obs}}, A\right),
\]
where, here, the auxiliary information $A$ denotes any user-specified information that conditions the completion, such as an intervention in a dynamical system, or a prompt in language modeling. Note that this is distinct from other conditional synthesis motivations previously discussed since the object of interest is not merely a static conditional $P(\cdot\mid A)$, but a trajectory-level conditional distribution that must respect temporal dependence.
Analysts typically retain $\mathcal O$ and use samples $\mathcal S\sim Q$ as completions conditional on observed partial trajectories. 

In missing-data settings, $\mathcal S$ provides plausible imputations of $Z_{\mathrm{miss}}$ given  $Z_{\mathrm{obs}}, A$) enabling downstream procedures that depend on a fully observed   dataset while propagating the uncertainty  due to the completion. This encompasses not only trajectory completion, but also the standard statistical missing-data problems where entries of $X$ and/or $Y$ are missing (e.g., unrecorded outcomes or incomplete covariates) and must be imputed for valid downstream inference. Modern generative approaches use flexible models to learn the conditional distribution of the missing components given the available data, producing augmented datasets that can be analyzed using complete data methods. 

For example, SynSurr \cite{mccaw2024synthetic} trains a model to generate outcomes using auxiliary information available for all individuals and then jointly analyzes the partially observed outcomes and full synthetic outcomes to yield unbiased regression coefficient estimates that are robust to the misspecification of generative models and increased power for downstream association studies (see Section \ref{sec:use}). Such  strategies are examples of synthetic data usage for data completion, where such synthetic data augment the observed data to support valid and more efficient statistical inference, relative to only relying on partially observed records.

In forecasting, namely the special case of $\mathcal I=\{1,\dots,t_0\}$, $\mathcal S$ provides plausible future continuations $Z_{t_0+1:T}\mid Z_{1:t_0}$ \cite{makarov2025large}. The conditioning variable $U$ also naturally supports counterfactual trajectory generation when it encodes an intervention \cite{melnychuk2022causal, luedtke2025doublegen}, and it accommodates prompt-conditioned completion in discrete sequence domains where the trajectory is a token string and $U$ is a textual prompt.

Several generators have been developed to explicitly model conditional dynamics and uncertainty in the presence of missingness. BRITS is a canonical deep-learning imputation approach that uses bidirectional recurrent modeling to infer missing values in time series from both past and future context \cite{cao2018brits}. Probabilistic imputation via score-based diffusion has also been implemented in approaches like CSDI \cite{tashiro2021csdi}. For trajectory synthesis, TimeGAN \cite{yoon2019time} provides a widely used template that couples adversarial training with losses designed to preserve time-series dynamics. In biomedical settings, emerging ``digital twin'' formulations frame trajectory completion and forecasting through large generative models that condition on patient history and potential interventions to generate forward trajectories, illustrating how this motivation extends beyond classical statistical imputation into conditional simulation for critical decision making \cite{elgammal2025digital}.

\subsection{Generative models for synthetic data}\label{subsec: generative model}

Recent advances in deep generative modeling have substantially improved sample fidelity and broadened the scope of synthetic data use. In this section, we briefly survey generative model families, without describing architectural or training details, for which comprehensive reviews exist (e.g., see \cite{bond2021deep}). 
Across these approaches, the statistical objective remains fitting an estimated distribution whose sampling law supports a particular synthetic data target $Q$ (Section \ref{sec:whySynth}). Generative model families differ in whether they specify an explicit likelihood $p_\theta(\cdot)$ or primarily define an implicit sampler without tractable densities. 

\begin{table*}[]
\scriptsize
\caption{Comparison of deep generative model classes}\label{tab:genmodels}
\setlength{\tabcolsep}{3pt}
\begin{tabular}{@{}m{2.5cm} m{3.5cm} m{3.5cm} m{7.5cm}}
\toprule
\multicolumn{1}{c}{\textbf{Model Class}} &
\multicolumn{1}{c}{\textbf{Statistical Object(s)}} &
\multicolumn{1}{c}{\textbf{Core Idea}} &
\multicolumn{1}{c}{\textbf{Strengths and Limitations}} \\ \midrule

\specialcell[b]{\textbf{Generative} \\\textbf{Adversarial Network} \\ \textbf{(GAN)}} &
Generator $G_{\theta_1}$ and discriminator $D_{\theta_2}$ (trained via adversarial loss) &
Train a generator to produce samples indistinguishable from real data by a discriminator adversary. &
\textit{Strengths:} High sample fidelity, effective at matching data distribution support without explicitly modeling likelihood\vfill \textit{Limitations:} Training can be unstable, susceptible to mode collapse where diversity of outputs is limited. \\ \midrule

\specialcell[b]{\textbf{Variational} \\ \textbf{Autoencoder (VAE)}} &
\specialcell[b]{
Encoder $q_{\theta_1}(l\mid z)$, decoder \\$p_{\theta_2}(z\mid l)$; ELBO objective \\$\mathbb{E}_{q_{\theta_1}(l\mid z)}[\log p_{\theta_2}(z\mid l)]-$ \\$\mathrm{KL}(q_{\theta_1}(l\mid z)\|p(l))$} &
Encoder-decoder architecture; learn latent variable representation with tractable approximate inference via a lower bound (ELBO). &
\textit{Strengths:} Interpretable latent representations, principled probabilistic model, stable training\vfill
\textit{Limitations:} Sample quality often low-resolution due to surrogate loss, posterior collapse. \\ \midrule

\textbf{Normalizing Flows} &
Invertible transform $f_\theta$; exact likelihood with $\log p_\theta(z)$ via Jacobian determinant &
Map a simple base distribution (e.g., Gaussian) through invertible functions to match data distribution &
\textit{Strengths:} Exact likelihood, smooth density modeling, invertible inference and sampling\vfill
\textit{Limitations:} Architecture can be constrained for tractability, challenging with discrete and high-dimensional data \\ \midrule

\specialcell[b]{\textbf{Autoregressive /} \\ \textbf{Transformer models}} &
Factorized likelihood: $p_\theta(z) = \prod_i p_\theta(z_i\mid z_{<i})$, attention weight matrices in transformer &
Model joint distribution sequentially using conditional factors; transformers use self-attention to capture long-range dependencies. &
\textit{Strengths:} Tractable likelihood, excellent for conditional generation and prompt-based synthesis.\vfill
\textit{Limitations:} Sequential sampling is inherently slower, often requires huge data \\ \midrule

\specialcell[b]{\textbf{Diffusion /}\\ \textbf{Score-based models}} &
Score function $s_\theta(z_t,t)\approx \nabla_z\log p_t(z_t)$ or reparameterization ($\varepsilon_\theta$, $z_\theta$); reverse SDE or denoising objective &
Forward noising process followed by learned reverse denoising; data generated by iterative refinement &
\textit{Strengths:} State-of-the-art sample fidelity and diversity, stable training, natural conditional extensions\vfill
\textit{Limitations:} Expensive iterative refinement during sampling \\ \bottomrule

\end{tabular}
\end{table*}

To fix notation, let $Z$ denote a generic data record, and let $\theta$ index a learned generative procedure; when an explicit density (or mass function) is available, we write it as $p_\theta(z)$.  Table \ref{tab:genmodels} summarizes five dominant deep generative model classes; for each class, the table identifies the relevant statistical objects, the core idea, and the practical tradeoffs that influence suitability across modalities and downstream protocols.

Classical deep generators remain important baselines and are often used when their assumptions align with the data modality or when a particular internal representation is useful downstream. {\it Generative adversarial networks} (GAN) \cite{goodfellow2014generative} learn a generator $G_{\theta_1}$ by an adversarial game against a discriminator $D_{\theta_2}$. Later frameworks interpret this training as divergence minimization (e.g., $f$-GAN \cite{nowozin2016f}) and proposing alternative objectives to improve stability (e.g., Wasserstein GAN \cite{arjovsky2017wasserstein}). While GANs typically produce sharp samples, well-known challenges include training instability and mode collapse \cite{goodfellow2014distinguishability, salimans2016improved, mescheder2018training, lin2018pacgan}. 
{\it Variational autoencoders (VAE)} \cite{kingma2013auto} posit a joint model $p_\theta(z,l)=p(l) p_\theta(z\mid l)$ with latent variable $L$, and fit $\theta$ by maximizing an evidence lower bound (ELBO) using an amortized approximation $q_\theta(l\mid z)$ to the intractable posterior $p_\theta(l\mid z)$. This yields a principled probabilistic model and a structured latent representation, but ELBO-based training can produce blurrier samples and can suffer from posterior collapse in expressive decoders. {\it Normalizing flows} \cite{rezende2015variational} also provide an explicit likelihood, but via an invertible change of variables. An invertible map $f_\theta$ is estimated from a tractable base distribution, enabling exact likelihood evaluation and exact latent inference, but architectural invertibility constraints can limit flexibility in high-dimensional or discrete settings.

For modern high dimensional data (e.g., for text and multimodal settings), autoregressive models and diffusion models have become central. {\it Autoregressive models} \cite{bengio2003neural} provide an alternative route to tractable likelihoods by factorizing the joint distribution over an ordered sequence representation $z=(z_1,\dots,z_d)$, namely $p_\theta(z)=\prod_{i=1}^d p_\theta(z_i\mid z_{<i})$, with classic implementations like PixelRNN image models \cite{van2016pixel} and WaveNet for audio \cite{oord2016wavenet}. 

The introduction of neural attention mechanisms has increased the adoption of autoregressive models as they provide a flexible way to mediate how each step in the generative process selectively references earlier sequence components. Early attention-based models exploited this idea to autoregressively construct complex objects, such as incrementally generating images by allowing the model at each step to decide where to ``attend to'' within previously generated content \cite{gregor2015draw}. 

The more recent development of self-attention in Transformer architectures \cite{vaswani2017attention}, has dramatically broadened the scope and effectiveness of autoregressive generative modeling. Self-attention replaces fixed sequential state updates with a parallel mechanism in which each position computes weighted interactions with all prior positions through an independent processing pathway. 

Concretely, embedded inputs are encoded as key-value pairs, where the values $\mathbf V$ encode the input content, and the keys $\mathbf K$ serve as an indexing structure. A query $q_i$ is made at each time step (stacking these queries across positions yields the query matrix $\mathbf Q$), and similarity scores are computed via dot products between the query and keys,
producing a weighting that determines how value vectors are aggregated. This can be expressed as:
\[
\mathrm{Attention}(\mathbf Q, \mathbf K, \mathbf V) = \mathrm{softmax}\left(\frac{\mathbf Q\mathbf K^\top}{\sqrt{d_k}}\right)\mathbf V,
\]
where $d_k$, the key/query dimension, is used to normalize gradient magnitudes. This self-attention mechanism enables stable training and efficient learning of long-range dependencies, which is particularly important at scale. The ability to model complex dependence structures without imposing strong Markovian constraints has led to dramatic empirical gains across text, image, and audio generation. Conditional generation can also be naturally implemented by conditioning the factorization on additional information, such as prompts. The main practical limitation is computational as sampling is inherently sequential in the token index.

Other generative model families learn a sampler through a stochastic generation process rather than a closed-form likelihood. {\it Diffusion  models}  \cite{ho2020denoising, sohl2015deep, song2019generative} 
define generation through a time-indexed corruption and denoising construction.  In the discrete-time DDPM formulation, a forward noising process gradually perturbs data $z_0\sim p_{\mathrm{data}}$ by
\[
q(z_t\mid z_{t-1})=\mathcal N\left(\sqrt{1-\beta_t}\,z_{t-1},\,\beta_t I\right),\quad t=1,\dots,T,
\]
for a chosen noise schedule $\{\beta_t\}$. This implies a closed-form marginal
\[
q(z_t\mid z_0)=\mathcal N\left(\sqrt{\bar\alpha_t}z_0,(1-\bar\alpha_t)I\right),\quad
\bar\alpha_t=\prod_{s=1}^t (1-\beta_s),
\]
so that samples at time $t$ can be generated as $z_t=\sqrt{\bar\alpha_t}\,z_0+\sqrt{1-\bar\alpha_t}\,\varepsilon$ with $\varepsilon\sim\mathcal N(0,I)$. The reverse-time generative mechanism then aims to sample from $p_\theta(z_{t-1}\mid z_t)$, implemented by a learned denoiser and an iterative refinement procedure. Training can be expressed via a simple denoising objective in which a network $\varepsilon_\theta(z_t,t)$ predicts the injected noise $\varepsilon$, or equivalently predicts $z_0$ or the score, yielding high-quality image synthesis with comparatively stable optimization.

Underlying these constructions is the Stein's score $\nabla_z \log p_t(z)$, namely the gradient field of the log-density of the perturbed data distribution at noise level $t$. In continuous time, the diffusion process can be represented as a forward SDE $dz=f(z,t)\,dt+g(t)\,d w_t$, where $w_t$ is a standard Wiener process (Brownian motion), $f$ is a drift term, and $g(t)$ controls the noise magnitude. Under mild regularity conditions, the corresponding reverse-time SDE depends explicitly on the score:
\[
dz=\left[f(z,t)-g(t)^2\nabla_z \log p_t(z)\right]\,dt + g(t)\,d\bar w_t,
\]
where $\bar w_t$ denotes a Weiner process with respect to the backward filtration generated by the diffusion process. This makes explicit that the reverse-time dynamics and sampling depend directly on on the score, and hence, learning a score network $s_\theta(z_t,t) \approx \nabla_z \log p_t(z)$ (or an equivalent parameterization) is precisely what makes generative sampling feasible in diffusion models \cite{song2020score, ho2020denoising}. 

This viewpoint also yields a deterministic probability flow ODE, a different sampler with the same marginals, whose drift again depends on the score, reinforcing the idea of estimating a time-indexed gradient field of log-densities. This connection between denoising and score estimation is a classical theme in denoising score-matching theory. The key practical tradeoff (Table~\ref{tab:genmodels}) is computational as sampling typically requires many reverse-time steps (SDE/ODE solver evaluations), though many improvements have been introduced to facilitate more efficient sampling \cite{song2020denoising, luo2023latent}.

Several relaxations of this typical diffusion construction have also been introduced yielding great success. Many approaches retain the continuous-time transport perspective but change the learned object from a score to a time-dependent velocity field. Flow-matching formulations learn a vector field $v_\theta(z,t)$ to transport a simple base distribution to the data distribution along a prescribed probability path, and then generate samples by solving the ODE $dz/dt=v_\theta(z,t)$ \cite{lipman2022flow}. Rectified-flow variants specialize the choice of probability path and training targets to encourage simpler transport trajectories for more efficient sampling \cite{liu2022flow}. These approaches primarily modify the statistical object being learned while preserving the same end goal, namely a high-fidelity sampler which can be easily leveraged to approximate the synthetic targets $Q$ in Section \ref{sec:whySynth}.

\subsubsection{Generative model choice.}

By construction, many generative model families can be applied to a given modality or task, though practical suitability depends on the downstream objective and constraints. Autoregressive models are a natural fit for text and audio generation due to their sequential factorization \cite{brown2020language, oord2016wavenet}; VAEs are attractive when downstream analysis benefits from a structured low-dimensional representation, though posterior collapse can arise with expressive decoders \cite{lucas2019don, razavi2019preventing}; and GANs can be parameter-efficient but may exhibit instability and can be less convenient for discrete data \cite{kumar2019melgan}. The choice of model class is also best determined by the downstream motivation. Since log-likelihood and sample fidelity can be weakly aligned in high dimensions \cite{theis2015note}, model choice is often guided by diagnostics that reflect the intended synthetic target $Q$ and use protocol (Table~\ref{tab:SDGmotivations}), such as conditional fidelity in data augmentation.

Conditional generation is key for linking generative model choice to downstream objectives. Conditional generation is often essential for the synthetic targets discussed in Section \ref{sec:whySynth} and is available across all model families in Table \ref{tab:genmodels}. Specifically, conditioning specifies a conditional target sampling law, where the conditional information may encode labels, attributes, domains, or prompts. Canonical examples include conditional VAEs \cite{sohn2015learning}, conditional GANs \cite{mirza2014conditional}, conditional autoregressive models such as conditional PixelCNN \cite{van2016conditional}, conditional diffusion with classifier-free guidance \cite{ho2022classifier}, and conditional flows \cite{winkler2019learning, trippe2018conditional}.

Beyond sampling considerations, the quality and utility of the internal representations learned during model fitting is a critical factor in generative model choice. For several model classes, learning the data-generating process also induces structured representations that can be directly exploited for downstream inference, even when synthetic sampling is not the primary objective. 
In semi-supervised learning, for example, Kingma et al. \cite{kingma2014semi} describe a latent-feature approach in which a generative model is used to learn an embedding from abundant unlabeled data, and a separate discriminative model is trained on this learned representation to improve classification. In genomics, approaches like scVI \cite{lopez2018deep} have been introduced which leverage representations learned from trained generative models to improve critical single-cell analysis tasks such as batch correction, visualization, clustering, and differential expression. 

Learned representations also enable information sharing across related datasets or institutions, which is especially valuable when local cohorts are small or labels are rare. This idea is closely related to multitask learning, where multiple related tasks or datasets are assumed to share an underlying representation, and joint learning is used to reduce sample complexity relative to fitting each task in isolation \cite{caruana1997multitask, evgeniou2004regularized, ando2005framework, bengio2013representation}. This common shared-structure assumption provides statistical value for learning a data generating process even when the goal is not necessarily sampling synthetic data.

\subsubsection{Modern extensions}
The scalability of generative models is an important consideration in modern high-dimensional settings. Recent approaches have turned to hybrid constructions that modify the learned object or the representation space while preserving the same target distributional goal. Latent diffusion models (LDM), for example, trains a diffusion model in a learned latent space (e.g., by first training a VAE and freezing weights before diffusion model training); this shifts computational burden from high-dimensional sampling to a lower-dimensional generative problem \cite{rombach2022high}.

In the original LDM formulation, the denoising component was implemented with a U-Net model \cite{ronneberger2015u}, which uses a contracting an expanding convolutional architecture with skip connections to efficiently model the denoising steps. Recently, diffusion transformer (DiT) models have also been proposed to replace convolutional denoisers with self-attention architectures that leverage transformer scaling behavior while retaining the denoising formulation of diffusion sampling \cite{peebles2023scalable}. Such generative procedures have shown improved performance in modern applications, namely, improved sample quality (e.g., lower Frechet Inception Distance on image generation benchmarks) and better scalability properties as model size and compute increase \cite{peebles2023scalable}. 
  
More recently, stochastic interpolant transformer (SiT) models have been proposed to unify diffusion and flow-based formulations within a transformer backbone. Leveraging interpolant-based training objectives, SiTs have been empirically shown state-of-the-art (SOTA) performance by improving sample quality and scalability (e.g., lower FID at comparable or reduced compute) \cite{ma2024sit}.

Modern applications increasingly require generative models capable of integrating diverse data types such as text, images, audio, and video. For example, multimodal large language models (MLLMs) extend autoregressive transformer backbones with modality-specific encoders and fusion mechanisms to jointly model and generate across modalities. By treating all data as sequences of tokens driven by a unified language modeling objective, these models enable coherent cross-modal generation and reasoning, as seen in prominent vision-language models like LLAVA \cite{liu2023visual}, Qwen-VL \cite{bai2023qwen}, and Gemini \cite{team2023gemini} that align pretrained vision encoders with text generation capabilities. 

Diffusion-based generative models have also been adapted for multimodal generation. Recent SOTA models move beyond generic conditioning toward architectures that explicitly couple modalities through shared attention or joint training objectives. Multimodal diffusion transformers (MM-DiT), for example, replace U-Net conditioning with deep cross-attention blocks or joint self-attention over concatenated modality tokens, allowing text, image, audio, or temporal streams to interact at every denoising step rather than through a shallow conditioning signal \cite{esser2024scaling}. Other hybrid frameworks have been proposed adapting pretrained text LLMs with parallel diffusion components to support interleaved text-image generation efficiently \cite{shi2024lmfusion}, while latent approaches unify continuous and discrete modalities in a shared generative process for multimodal modeling \cite{sun2024multimodal}.

\section{The use of synthetic data in downstream analysis}
\label{sec:use}
\subsection{The use of synthetic data: model misspecification, uncertainty, efficiency, and generalization ability}
Beyond the generation of synthetic data, a central methodological question concerns how such data should be used in downstream tasks, including parameter estimation and prediction. While synthetic data are increasingly employed to mitigate data scarcity and distributional shifts,  their effective use to ensure valid statistical inference and model generalization ability is nontrivial because synthetic observations are generated rather than directly observed, and generative models may be misspecified.

Three interrelated challenges arise when incorporating synthetic data into downstream analysis. First, synthetic data inherit biases from the underlying generative model. Misspecification of the generative mechanism may distort marginal distributions, dependence structures, or tail behavior, leading to biased estimators and unreliable predictions. Second, uncertainty introduced during the synthetic data generation process must be properly  taken into account in downstream statistical analysis. Treating synthetic data as fixed or equivalent to real observations often results in underestimation of uncertainty and invalid inference. Third, even when synthetic data are informative,  carefully developed statistical methods are required  to efficiently combine them with real observed data to balance robustness, efficiency gains, and model generalization ability.

These considerations have led to several distinct paradigms for using synthetic data, each making different assumptions and offering different trade-offs among validity, robustness, efficiency, and generalization ability. 

To fix the idea, we consider the problem of the regression problem  $Y=f(X; \beta)+\epsilon$ as a running example in the following sections. Specifically, at the population level, our goal is to minimize the prediction loss $E_{P_{\cT}}[\{Y - f(X; \beta)\}^{2}]$ on the target population. We assume all variables are mean zero for the ease of illustration.

\subsection{Synthetic data-based approaches}\label{subsec: based}
The most direct paradigm generates synthetic data $\mathcal S$ to mimic the real data $\mathcal O$ and leverages the synthetic dataset $\mathcal S$ or the combined data $\mathcal O \cup \mathcal S$ to estimate parameters or train prediction models as if they were all real data, using standard estimation or learning procedures without modification \citep{an2023deep,zhu2024distribution}. In this setting, synthetic data are effectively viewed as real data in downstream tasks. 
  This paradigm is usually applied in cases where the target distribution aligns with the training distribution, i.e., $P_{\cT} = P$. In the above regression problem, DistDiff \citep{zhu2024distribution} generates synthetic data $\mathcal S = \{(\hX_{i}, \hY_{i})\}_{i = 1}^{N}$ using a distribution-aware diffusion model that approximates the training data distribution, and trains the regression model by minimizing
\[
\sum_{i = 1}^{n}\{Y_{i} - f(X_{i}; \beta)\}^{2} + \sum_{i = 1}^{N}\{\hY_{i} - f(\hX_{i}; \beta)\}^{2}
\]
with respect to $\beta$. 

AutoComplete \citep{an2023deep} considers a semi-supervised regression problem where the label $Y$ is observed for a subset of $\cO$. Suppose $\cO = \{(X_{i}, Y_{i})\}_{i = 1}^{n_{\rm obs}}\cup \{X_{i}\}_{i = n_{\rm obs} + 1}^{n}$ where $n_{\rm obs}$ is the number of labeled data. AutoComplete first generates synthetic labels $\{\hY_{i}\}_{i = n_{\rm obs} + 1}^{n}$ for the unlabeled data based on neural networks and available information of the unlabeled units (e.g., observed surrogates of the label or informative covariates). Then, the AutoComplete (AC) estimator $\hat{\beta}_{\scaleto{\rm AC}{3.5pt}}$ for $\beta_{\star}$ is obtained by minimizing
\[
\sum_{i = 1}^{n_{\rm obs}}\{Y_{i} - f(X_{i}; \beta)\}^{2} + \sum_{i = n_{\rm obs} + 1}^{n}\{\hY_{i} - f(X_{i}; \beta)\}^{2}.
\]

The appeal of synthetic data-based approaches lies in their simplicity and scalability. It can improve efficiency by expanding the sample size. When the generative model is correctly specified and accurately estimated, synthetic data can increase the effective sample size and potentially improve convergence rates \citep{tian2025conditional}. In idealized settings, this paradigm may lead to substantial efficiency gains.

However, these benefits rely critically on the correct specification of generative models. Misspecified models can propagate systematic errors throughout the analysis, leading to biased estimators and/or misleading predictions \citep{kang2007demystifying,mccaw2024synthetic}. Moreover, uncertainty induced by the synthesis step is typically ignored \citep{an2023deep}, further undermining inferential validity \citep{drechsler202430}. A statistical look at the implications of model misspecification and uncertainty of the synthetic data can be essential for the safe use of synthetic data-based approaches.

\subsection{Synthetic data-assisted approaches}\label{subsec: assisted}
A more robust and principled paradigm uses synthetic data $\mathcal S$ that mimics the real data as an auxiliary resource while retaining real data $\mathcal O$ as the primary basis for identification and inference. 
Synthetic data-assisted approaches are suitable for the case where $P_{\cT} = P$.

 A direct way to use synthetic data to assist data analysis is to tune a method based on synthetic data, e.g., choosing the kernel to use in tests \citep{chang2019kernel}. In this case, the validity of the analysis results is guaranteed regardless of whether the generative model is correctly specified, provided that the analysis method is valid for any given kernel or tuning parameter. The synthetic data help to choose an appropriate kernel or tuning parameter and hence improves the performance of the method if the generative model closely mimics the real data. 

The synthetic data-assisted paradigm is widely adopted in semi-supervised inference, where synthetic data are generated based on unlabeled data and are treated as surrogates or nuisance components of the influence functions in semiparametric theory. Examples include Prediction-Powered Inference (PPI) \citep{angelopoulos2023prediction}, Synthetic Surrogate (SynSurr) \citep{mccaw2024synthetic} and several other methods
\citep{miao2024valid, ji2025predictions,lyu2025bias}. Specifically, 
in the semi-supervised regression problem introduced in Section \ref{subsec: based}, PPI \citep{angelopoulos2023prediction} generates synthetic labels $\{\hY_{i}\}_{i = 1}^{n}$ for both labeled and unlabeled data, and obtain the estimator $\hat{\beta}_{\scaleto{\rm PPI}{3.5pt}}$ 
by minimizing
\[
\begin{aligned}
    &\sum_{i = 1}^{n_{\rm obs}}\{Y_{i} - f(X_{i}; \beta)\}^{2} - \sum_{i = n_{\rm obs} + 1}^{n}\{\hY_{i} - f(X_{i}; \beta)\}^{2}\\
    &+ \frac{n_{\rm obs}}{n - n_{\rm obs}}\sum_{i = n_{\rm obs} + 1}^{n}\{\hY_{i} - f(X_{i}; \beta)\}^{2}.
\end{aligned}
\]
The above objective function scaled by a factor $n_{\rm obs}^{-1}$ is an unbiased estimate of the population loss $E_{P}[\{Y - f(X; \beta)\}^{2}]$ provided that $(\hY_{i}, X_{i})$ is identically distributed across $i = 1,\dots, n$ regardless whether the model that generates $\hY_{i}$ is correctly specified or not. Thus, the resulting estimator $\hat{\beta}_{\scaleto{\rm PPI}{3.5pt}}$ can be consistent and asymptotically normal even if the generative model is misspecified \citep{angelopoulos2023prediction}.   However, it requires a strong missing data assumption, i.e., the data are missing completely at random, and the estimator $\hat{\beta}_{\scaleto{\rm PPI}{3.5pt}}$ can be less efficient than the estimator using  only labeled data unless the generative model is highly predictive, i.e., it is subject to negative learning.

The Synthetic Surrogate (SynSurr) approach \citep{mccaw2024synthetic} is more robust and powerful. It considers the semisupervised linear regression where $f(X; \beta) = X\beta$ and obtain the estimator $\hat{\beta}_{\scaleto{\rm SynSurr}{3.5pt}}$ 
by minimizing
\[
\begin{aligned}
    &\sum_{i = 1}^{n_{\rm obs}}(Y_{i} - X_{i}\beta - \hat{e}_{i}\alpha)^{2},
\end{aligned}
\]
where $\hat{e}_{i}$ is the regression residual of the regression $\hY \sim X$ based on $\{(X_{i}, \hY_{i})\}_{i = 1}^{n}$. Note that $\hat{e}_{i}$ is asymptotically orthogonal to $X_{i}$ in the sense that $E(\hat{e}_{i}X_{i}) \to 0$ under regularity conditions provided that $(\hY_{i}, X_{i})$ is identically distributed across $i = 1,\dots, n$. Thus, the regression coefficient of $X$ in the $Y \sim X$ regression and the $Y\sim X + \hat{e}$ regression converge to the same probability limit and are both asymptotically normal under regularity conditions. This justifies the validity of the inference based on SynSurr even when the generative model is misspecified. Moreover, SynSurr can achieve a significant improvement in efficiency in estimating $\beta$ over the regression using only the labeled data, if the generative model is predictive
and the proportion of labeled data is small \citep{mccaw2024synthetic}.  In addition, it requires a weaker assumption on the missing data mechanism, i.e., requiring the data are missing at random.

\begin{figure}[H]
    \centering
    \includegraphics[width=0.9\linewidth]{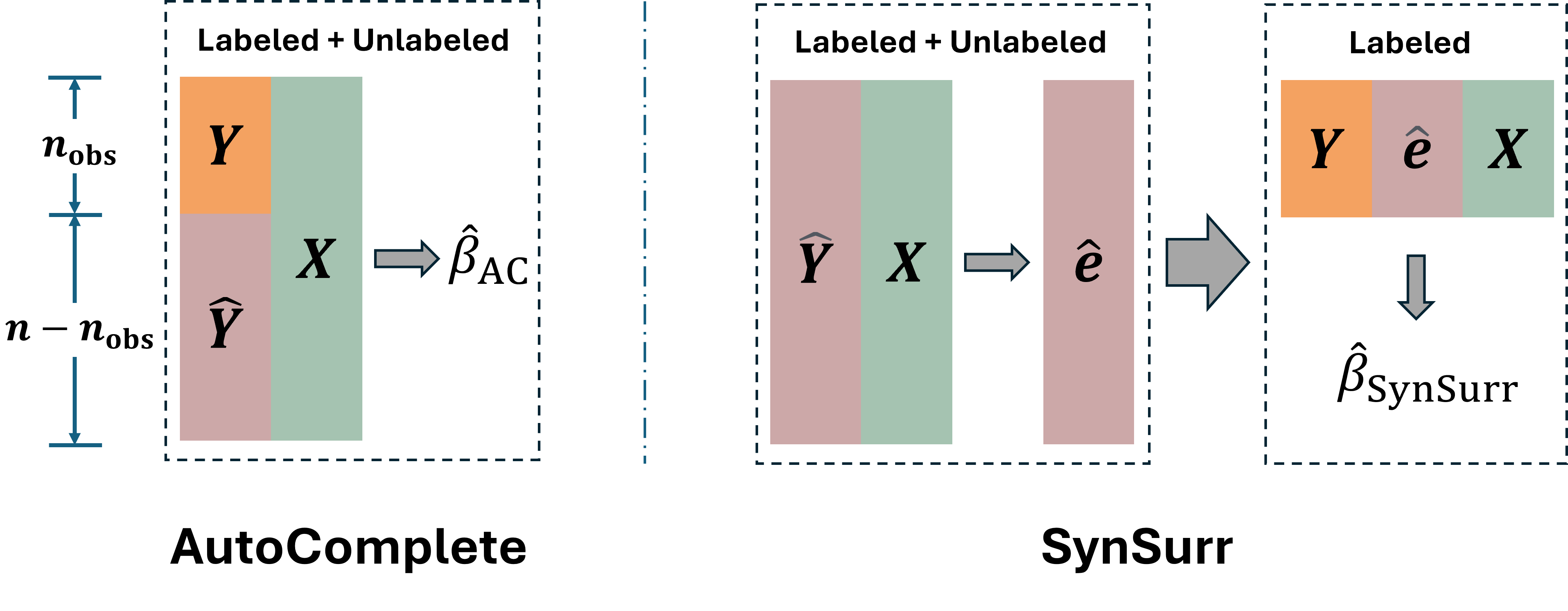}
    \caption{An illustration of the difference between synthetic data-based and synthetic data-assisted approaches in the semisupervised regression context using AutoComplete \citep{an2023deep} and SynSurr \citep{mccaw2024synthetic} as examples. AutoComplete pools the synthetic data and real data to conduct inference, while SynSurr uses synthetic data to construct $\hat{e}$ to assist the inference based on labeled data.}\label{fig: semi-supervise}
\end{figure}

Figure \ref{fig: semi-supervise} illustrates the difference between synthetic data-based and synthetic data-assisted approaches in the semisupervised regression context using AutoComplete \citep{an2023deep} and SynSurr \citep{mccaw2024synthetic} as examples.
A key advantage of synthetic data-assisted approaches is robustness. Such procedures usually inherit the validity of some real data-based statistical methods, such as kernel-based testing \citep{gretton2012kernel} and the augmented inverse probability weighted method \citep{tsiatis2006semiparametric}, even when the generative model is misspecified.

In synthetic data-assisted approaches, estimation and inference rely on real data, with synthetic data improving statistical efficiency.
 In estimation problems, these methods typically achieve the same asymptotic convergence rates as real-data-based estimators. The benefits of synthetic data are therefore primarily in reducing the asymptotic variance of regression coefficient estimators and improving estimation efficiency and test power. Compared to synthetic data-based approaches, the synthetic data-assisted paradigm 
 balances 
 validity/robustness 
 and statistical efficiency gain.

\subsection{Synthetic data-augmented approaches}\label{subsec: augmented}
A third line of work uses synthetic data to augment real datasets by generating perturbed samples that represent unseen, rare, or counterfactual scenarios \citep{zhang2018mixup,wang2022out,chen2024comprehensive,rahat2025data}. Rather than approximating the observed data distribution, these approaches explicitly aim to extrapolate beyond it, thereby increasing both the size and diversity of the available data.

Synthetic data augmentation is often motivated by concerns about spurious correlations, covariate shift, or insufficient coverage of the input space. By introducing carefully designed synthetic variations, these methods seek to improve robustness and out-of-distribution generalization. In this sense, synthetic data functions as a tool for stress-testing models and enriching the training signal.  Unlike synthetic data-based and synthetic data-assisted approaches, synthetic data augmented approaches are particularly valuable when the target distribution $P_{\cT}$ differs from the training distribution $P$.

In synthetic data-augmented approaches, synthetic data $\mathcal S$ are either directly involved in model training \citep{zhang2018mixup,zhoudomain,chen2024comprehensive} or utilized to regularize the model \citep{wang2022out} to enhance generalization ability.
 An example of method that directly involve synthetic data augmentation to model training is CoDSA \citep{tian2025conditional}. To address the data scarcity in certain regions, CoDSA splits the support of $P$ into $K$ different regions, generates synthetic data $\{(\hX_{ki}, \hY_{ki})\}_{i = 1}^{N_{k}}$ from the $k$-th region using a conditional generative model for $k = 1,\dots, K$ and a prespecified size $N_{k}$. Then, in the regression example, CoDSA trains the regression model by minimizing
\[
\sum_{i = 1}^{n}\{Y_{i} - f(X_{i}; \beta)\}^{2} + \sum_{k = 1}^{K}\sum_{i = 1}^{N_{k}}\{\hY_{ki} - f(\hX_{ki}; \beta)\}^{2}
\]
with respect to $\beta$. Clearly, the above objective function is targeted on a distribution different from the original training distribution $P$ with intentionally oversampling in certain regions (determined by the choice of $N_{1}, \dots, N_{K}$). 

Leveraging synthetic data, CoDSA can increase the sample size from specific data-scarce regions which can potentially improve the generalization ability of the trained model to distributions different from the training distribution.
Note that the loss function used in CoDSA closely resembles that of AutoComplete, as both approaches treat synthetic data as if they were real observations. As a result, CoDSA requires a correct specification of the generative model and is subject to bias in the presence of misspecification of the generative model.

\begin{figure}[H]
    \centering
    \includegraphics[width=0.9\linewidth]{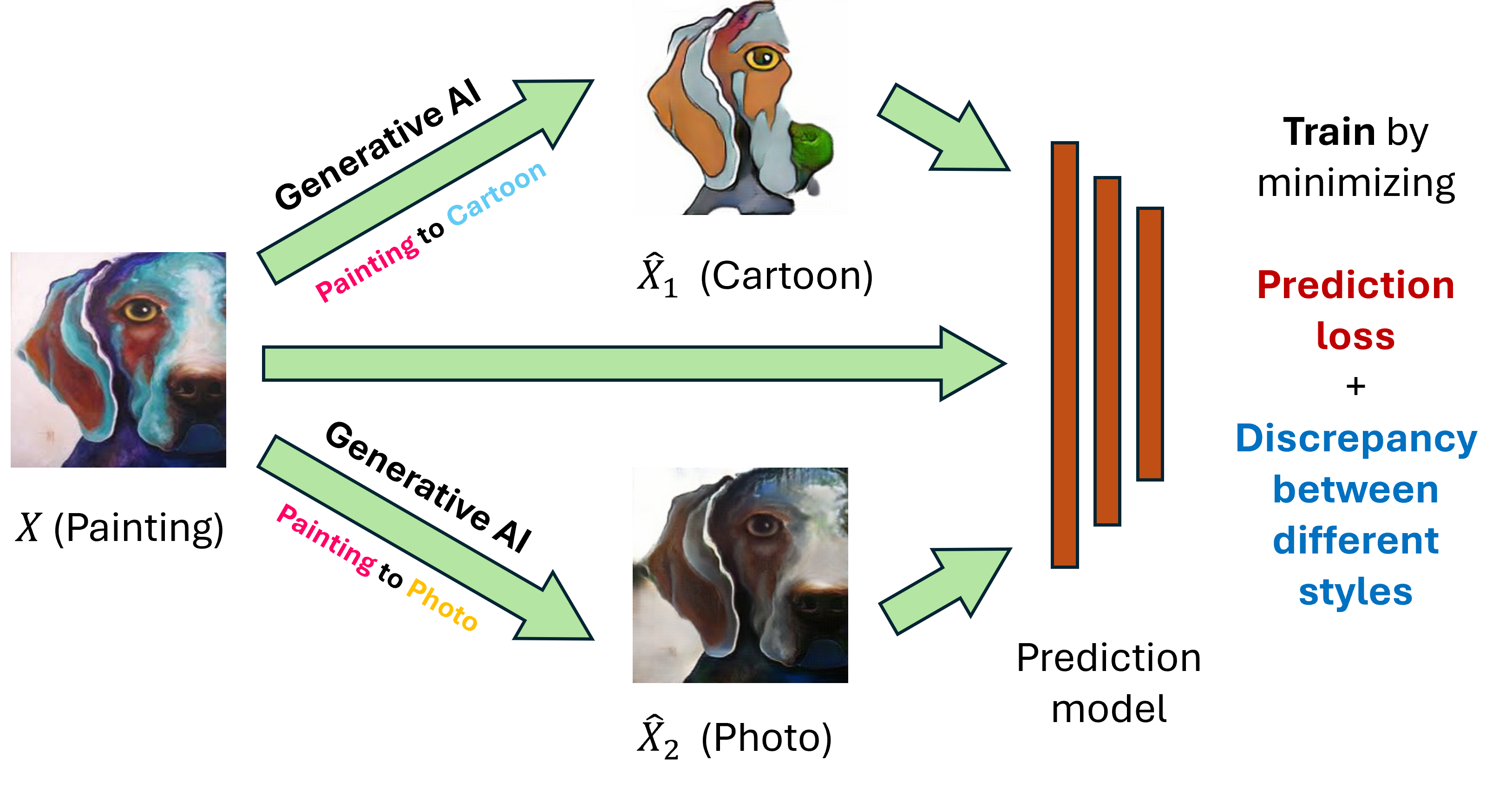}
    \caption{An illustration of RICE \cite{wang2022out} -- a regularization-based synthetic data-augmented method.   To build a model to robustly classify images of diverse styles, RICE generates synthetic images of different styles (e.g., cartoon and photo) based on each real image (e.g., painting). The RICE training procedure impose regularization terms to encourage the model to perform similarly on real images and their synthetic counterparts.  
}\label{fig: RICE}
\end{figure}

RICE \citep{wang2022out} is an example of the regularization-based synthetic data-augmented method. To introduce the idea of RICE, let us consider $X$ as an image and $Y$ as the label we want to predict based on $X$. In the real world, the image might be of different styles, e.g., painting, cartoon, photo, etc., but the inputs in the training data $\{(X_{i}, Y_{i})\}_{i = 1}^{n}$ may only consist of one style, say painting. To enhance the model's generalization ability to inputs of other styles, RICE transforms each original input $X_{i}$ to synthetic data of different styles $\{X_{ik}\}_{k = 1}^{K}$ using AI tools such as CycleGAN \citep{zhu2017unpaired} and add penalties to enforce the model to output similar results on real image and its synthetic counterpart. Specifically, RICE trains the model by minimizing
\[
\begin{aligned}
    \sum_{i = 1}^{n}\left[\{Y_{i} - f(X_{i}; \beta)\}^{2} + \lambda\max_{k} D\{f(\hX_{i}; \beta), f(\hX_{ik}; \beta)\}\right],
\end{aligned}
\]
where $\lambda$ is a tuning parameter and $D\{\cdot, \cdot\}$ is some measure of
discrepancy. The penalty in RICE encourages the model to use non-style-specific features to predict the label, which can improve the generalization ability of the model on inputs of different styles. Figure \ref{fig: RICE} provides an illustration of RICE.

Synthetic data-augmented approaches can improve predictive performance on target populations that differ from the training population. However, conducting valid statistical inference under data-augmented approaches remains challenging and largely open, due to the difficulty of characterizing both the randomness introduced by synthetic data and the errors arising from the generative model.
In addition, synthetic data augmentation-based approaches depend heavily on prior knowledge to determine which synthetic perturbations are meaningful. Poorly designed augmentations may introduce unrealistic patterns or amplify modeling bias. Consequently, their success hinges on domain-informed construction of the generative process. While practically useful, there is still a lack of recognized statistical frameworks to characterize the conditions or assumptions underpinning the synthetic data augmentation-based approaches to our knowledge.

\subsection{In-Context Learning Based on Synthetic Data}

Recent work has explored the use of synthetic data within in-context learning frameworks, particularly in large neural architectures such as transformers \citep{mullertransformers,hollmanntabpfn,hollmann2025accurate}. In contrast to conventional uses of synthetic data that aim to approximate a single target distribution, these approaches generate a large and diverse collection of synthetic tasks where each synthetic task consists of a synthetic training dataset $\mathcal{S}_{\rm train}^{(j)} = \{(\hX_{{\rm train}, i}^{(j)}, \hY_{{\rm train}, i}^{(j)})\}_{i = 1}^{N_{{\rm train}}^{(j)}}$ and a synthetic test dataset $\mathcal{S}_{\rm test}^{(j)} = \{(\hX_{{\rm test}, i}^{(j)}, \hY_{{\rm test}, i}^{(j)})\}_{i = 1}^{N_{{\rm test}}^{(j)}}$ for $j = 1, \dots, M$. Instead of training a model to predict the outcome $Y$ based on $X$, in context learning aims to train a model $q(\cdot\mid \cdot, \cdot; \phi)$ that takes a training dataset and the predictors in a test dataset as input, and produces predicted $X$ distributions for the corresponding labels $Y$ in the test dataset, where $\phi$ is the model parameter. The objective is not direct prediction for a given target distribution, but rather to enable the model to learn which prediction strategy should be used for each dataset. Training proceeds by maximizing the average test likelihood
\[
\frac{1}{M}\sum_{j = 1}^{M}\log q(\widehat{\mathbb Y}_{\rm test}^{(j)}\mid (\widehat{\mathbb X}_{\rm test}^{(j)}, {\mathcal S}^{(j)}; \phi)
\]
across all synthetic tasks, where 
\[\widehat{\mathbb X}_{\rm test}^{(j)} = \left(\hX_{{\rm test}, 1}^{(j)}, \dots, \hX_{{\rm test}, N_{{\rm test}}^{(j)}}^{(j)}\right)^{\T}\]
and
\[
\widehat{\mathbb Y}_{\rm test}^{(j)} = \left(\hY_{{\rm test}, 1}^{(j)}, \dots, \hY_{{\rm test}, N_{{\rm test}}^{(j)}}^{(j)}\right)^{\T}.\]
Let $\hat{\phi}$ be the model parameter trained based on the synthetic data. When applied to a real-world problem with training data $\mathcal O_{\rm train}$ and test data predictors ${\mathbb X}_{\rm test} = (X_{{\rm test}, 1}, \dots, X_{{\rm test}, n_{\rm test}})^{\T}$, the trained model directly produces a predicted distribution $q(\cdot\mid {\mathbb X}_{\rm test}, {\mathcal O}_{\rm train}; \hat{\phi})$ for the labels corresponding to ${\mathbb X}_{\rm test}$ without any parameter updates or fine-tuning. 

Intuitively, by exposing to diverse synthetic tasks, the model can learn how to adapt to different data distributions and automatically select the "method" and "tuning parameter" to use based on the real training data $\mathcal O_{\rm train}$ \citep{bai2023transformers}. Thus, no manual adaptation is required.
From a statistical perspective, synthetic tasks serve as an implicit prior over tasks or data-generating processes. By training on a broad family of synthetic tasks, models can learn to perform approximate Bayesian prediction, adapt to new datasets, and exhibit strong zero-shot performance. The success of these methods \citep{hollmann2025accurate} suggests that, under appropriate task generation schemes, synthetic data can encode rich structural information about statistical problems.

However, this paradigm raises several methodological challenges. First, the effectiveness of in-context learning depends critically on the realism and diversity of the synthetic tasks. If the synthetic task distribution poorly reflects real-world data-generating mechanisms, the learned prediction strategies may fail to transfer \citep{wang2025can}. Second, unlike traditional statistical procedures, theoretical guarantees regarding consistency, efficiency, or robustness remain limited, making it difficult to assess reliability in high-stakes applications. Building a statistical understanding of the conditions under which synthetic task generation leads to reliable and transferable prediction results is an open problem. Third, in-context learning based on synthetic data mainly focuses on prediction problems. Developing in-context learning methods tailored for inference and uncertainty quantification remains an important open direction.

\begin{table}[h]
\centering
\caption{Comparison of different paradigms for using synthetic data in downstream tasks.}
\label{tab:synthetic_data_paradigms}
\resizebox{0.45\textwidth}{!}{
\begin{tabular}{m{1.5cm}<{\raggedright} m{1.3cm}<{\raggedright} m{1.3cm}<{\raggedright} m{2cm}<{\raggedright} m{4cm}<{\raggedright}}
\hline
\textbf{Paradigm} 
& \textbf{Role of Synthetic Data} 
& \textbf{Target Distribution} 
& \textbf{Convergence Rate} 
& \textbf{Main Strengths and Limitations} \\
\hline
Synthetic data--based 
& Treated as real data 
& $P_{\cT} = P$
& Potentially improved if the model is correct 
& \textit{Strengths:} simple and scalable \vfill
\textit{Limitations:} highly sensitive to generative model error and ignores synthesis uncertainty \\
\hline
Synthetic data--assisted 
& Auxiliary to real data 
& $P_{\cT} = P$ 
& Same as real data-based methods (asymptotic variance can be improved)
& \textit{Strengths:} robust to generative model misspecification;\vfill 
\textit{Limitations:} efficiency gains are usually limited to a constant level, no rate improvement \\
\hline
Synthetic data--augmented 
& Generate unseen or rare samples 
& $P_{\cT} \neq P$
& Unclear 
& \textit{Strengths:} can improve generalization and robustness to unseen heterogeneous target data\vfill 
Limitation: relies on strong prior knowledge for realistic augmentation\\
\hline
Synthetic in-context-learning 
& Generate synthetic tasks 
& $P_{\cT} \neq P$
& Unclear
& \textit{Strengths:} learns general prediction and inference strategies across tasks and can automatically adapt to different problems without training\vfill 
\textit{Limitations:} adaptivity and transferability depend on task realism \\
\bottomrule
\end{tabular}
}
\end{table}

\section{Other topics on the use of synthetic data and future directions}






\smallskip
{\noindent\bf Generation of High-fidelity Data Suitable for Downstream Tasks}
\smallskip

Although many advanced AI techniques for generating synthetic data are available, as reviewed in Section~\ref{subsec: generative model}, developing generative models that produce high-fidelity synthetic data suitable for downstream tasks remains an important and ongoing challenge \citep{lu2023machine}. High fidelity in this context goes beyond marginal distributional similarity and requires preservation of structural properties that are relevant to the target task, such as conditional relationships, dependence structures, tail behavior, or causal mechanisms \citep{kim2025technical,jiang2025tabstruct}. While modern generative models can produce visually or descriptively realistic samples, it remains difficult to assess whether they accurately capture the aspects of the data-generating process that matter for inference or prediction \citep{alaa2022faithful,kim2025technical}. Developing principled criteria for evaluating and improving the fidelity of synthetic data--particularly in a task-aware manner--remains an important open direction. 

\smallskip
{\noindent\bf Trade-offs among Validity, Robustness, and Efficiency When Integrating Synthetic and Real Data}
\smallskip

Beyond data generation, integrating synthetic and real data in a way that is both statistically valid and practically powerful presents additional challenges. Naively pooling synthetic and real observations can lead to biased inference if errors in synthetic data are not properly accounted for, while overly conservative approaches may fail to exploit the potential efficiency gains offered by synthetic data. As reviewed in Sections \ref{subsec: based} and \ref{subsec: assisted}, synthetic data-based and synthetic data-assisted paradigms represent two distinct trade-offs among validity, robustness, and efficiency. Developing additional paradigms that achieve alternative balances among these objectives would be valuable. Moreover, a key open question in this direction is how to design adaptive integration strategies that balance robustness and efficiency, allowing synthetic data to contribute information where they are reliable while limiting their influence where they are not. Future research will need to develop principled weighting, calibration, and diagnostic tools that govern effective integration of synthetic and real data in general settings. Addressing these challenges is essential for realizing the full potential of synthetic data in statistical inference and learning.

\smallskip
{\noindent\bf Uncertainty Propagation from Data Synthesis}
\smallskip

  Despite rapid progress, several fundamental challenges remain in the principled use of synthetic data for statistical inference. One of the most common challenges is the uncertainty introduced by the synthetic data generation process. Because synthetic data are produced by estimated generative models rather than directly observed, they carry additional sources of variability and bias that are fundamentally different from classical sampling noise.
Most existing methods, except for the synthetic data-assisted approaches reviewed in Section \ref{subsec: assisted}, ignore this synthesis-induced uncertainty. In synthetic data-based and augmented approaches, synthetic observations are often treated as fixed or equivalent to real data, which can lead to systematically misestimated uncertainty and invalid inference.  

Developing general inferential frameworks that explicitly account for synthesis uncertainty in downstream estimation and inference, therefore, remains an important open problem. 
An appealing direction to addressing this problem is to draw inspiration from existing inferential paradigms that can accommodate black-box machine learning algorithms, such as double machine learning \citep{chernozhukov2018double} and conformal inference \citep{shafer2008tutorial}, and extend these ideas to settings where synthetic data play a central role.

\smallskip
{\noindent\bf  Extrapolation, Generalization, and Transfer}
\smallskip

  As discussed in Section \ref{subsec: augmented},
synthetic data are increasingly used to improve out-of-distribution generalization through data augmentation and controlled extrapolation. By generating samples that cover underrepresented regions of the input space or simulate hypothetical shifts in the data-generating process, synthetic data can expose models to a wider range of scenarios than those in the training data. This strategy is particularly useful in settings involving distribution shift, where real training data may fail to represent the unknown target distribution adequately.

In the area of transfer learning \citep{wang2023pseudo,tian2025conditional}, synthetic data provides a flexible mechanism for bridging gaps between training and target distributions. Rather than relying on reweighting or feature alignment, synthetic data can be used to directly construct synthetic training distributions that mimic the target distribution. When the generative model captures structural invariances shared between populations, training based on synthetic augmentation holds promise to improve transfer performance.

Despite empirical successes, the theoretical understanding of synthetic-data-driven out-of-distribution generalization and transfer learning remains limited. In particular, the understanding of the conditions under which synthetic extrapolation can preserves the relevant causal or predictive relationships needed for successful generalization and transfer is still insufficient. Poorly specified generative models may introduce unrealistic samples, distort conditional relationships, or amplify spurious correlations, thereby degrading generalization rather than improving it. 
A central open problem is therefore to characterize when and how synthetic data improve generalization ability and transferability. This includes, but is not limited to, identifying the types of distributional shifts for which synthetic augmentation is beneficial, understanding the role of the estimation error of the generative model, and developing diagnostics to detect harmful extrapolation.

\smallskip
{\noindent\bf Privacy protection when using synthetic data}
\smallskip

Although synthetic data are often motivated by privacy concerns, releasing only synthetic data does not automatically guarantee privacy. In practice, generative models may memorize or leak sensitive information from the training data, especially when trained on small or highly structured datasets. Consequently, caution is required when synthetic data are used for privacy protection.

Several approaches have been developed to provide formal privacy guarantees, including methods based on differential privacy and privacy-preserving training mechanisms \citep{jordon2019pate,chourasia2021differential,tsai2025differentially}. These methods typically trade data utility for provable privacy protection, highlighting an inherent tension between fidelity and privacy. How to optimally balance the data utility and provable privacy protection is an important future direction for guiding the use and release of synthetic data.

\smallskip
{\noindent\bf Computational Considerations}
\smallskip

The use of synthetic data raises important computational questions. Generating high-fidelity synthetic data can be computationally expensive, particularly when training large generative models. Moreover, incorporating synthetic data into downstream analysis introduces additional design choices, such as how many synthetic samples to generate and how to weight them relative to real data.

These choices reflect a broader statistical--computational trade-off. Excessive synthetic data may increase computational burden without improving statistical accuracy, while insufficient synthetic data may fail to deliver practical benefits. Understanding how to optimally balance data quality, quantity, and computational cost largely remains an open and practically relevant problem.

\smallskip
{\noindent\bf  Understanding In-Context Learning from Synthetic Data}
\smallskip

A particularly important open problem concerns the theoretical and methodological understanding of in-context learning frameworks trained on synthetic data. While empirical evidence suggests that such models can internalize effective inference strategies through exposure to synthetic task distributions, it remains unclear what aspects of the synthetic task generation process are essential for successful transfer to real-world data. Key questions include how the choice of task family influences learned inductive biases, under what conditions in-context inference approximates optimal or Bayesian procedures, and how robustness to task misspecification can be ensured.

Moreover, unlike traditional estimators, in-context learning systems embed inference procedures implicitly within model parameters, complicating uncertainty quantification, interpretability, and formal guarantees. Establishing connections between synthetic-task-based in-context learning and classical statistical concepts--such as sufficiency, efficiency, robustness, and Bayesian inference--represents an important direction for future research.

\section*{ACKNOWLEDGMENT}

This work was supported by grants from the National Institute of Health R35-CA197449, R01-HL163560, U01-HG012064, U19-CA203654, and P30 ES000002.

\bibliographystyle{imsart-nameyear}
\bibliography{Syn}
\end{document}